\documentclass[runningheads]{llncs}

\usepackage{lipsum}
\usepackage[T1]{fontenc}

\usepackage{graphicx}

\usepackage[ruled,vlined]{algorithm2e}
\usepackage{amsmath}
\usepackage{amsfonts}
\usepackage{amssymb}

\begin{document}
\title{\textsc{DP-IVON-Gradsq}: Differentially Private Squared-Gradient \\ Improved Variational Online Newton}
\titlerunning{\textsc{DP-IVON-Gradsq}}

\author{
Nour Jamoussi\inst{1}\thanks{Corresponding author.} \and
Ikram Dridi\inst{2} \and
Giuseppe Serra\inst{1} \and
Marios Kountouris\inst{1,3}
}

\authorrunning{Jamoussi et al.}

\institute{
EURECOM, Sophia Antipolis, France\\
\email{nour.jamoussi@eurecom.fr}
\and
Concordia University, Montreal, Canada
\and
University of Granada, Granada, Spain \\
\email{mariosk@ugr.es}
}

\maketitle            
\begin{abstract}
Differential privacy provides formal privacy guarantees for training neural networks on sensitive data, while Bayesian deep learning offers a principled framework for uncertainty-aware prediction. Combining these two objectives remains challenging, as privacy noise can interact with the stochasticity introduced by Bayesian posterior sampling. In this work, we investigate differentially private variational Bayesian learning through the Improved Variational Online Newton (IVON) optimizer. We introduce \textsc{DP-IVON-Gradsq}, a private variant of IVON. The proposed method constructs its curvature estimate from the privatized gradient using a noise-corrected squared-gradient estimator, reducing the direct interaction between posterior-sampling noise and privacy noise while preserving the Adam-like computational efficiency of IVON. We evaluate \textsc{DP-IVON-Gradsq} on CIFAR-10 against the standard private optimizers DP-SGD and DP-Adam over a range of privacy budgets. The results show that \textsc{DP-IVON-Gradsq} is competitive under weak-to-moderate privacy constraints, i.e., large-to-moderate values of $\varepsilon$, while degrading under strong privacy. Code is available at \url{https://github.com/NourJamoussi/DP-IVON-Gradsq.git}.

\end{abstract}
\section{Introduction}

Deep neural networks are increasingly deployed in sensitive domains such as healthcare, finance, and mobile applications, where training data may contain personal and confidential information. This raises important privacy concerns, reinforced by data-protection regulations such as the General Data Protection Regulation (GDPR), which require careful handling of personal data throughout the machine learning pipeline. Nevertheless, many high-stakes applications require models not only to make accurate predictions, but also to quantify uncertainty, especially under limited data or distribution shift. Bayesian deep learning provides a principled framework for uncertainty-aware prediction by learning a distribution over model parameters instead of learning a single deterministic solution. However, when Bayesian models are trained on sensitive datasets, the learned posterior may still reveal information about individual training examples. This creates a need for learning algorithms that combine the uncertainty-awareness of Bayesian deep learning with formal privacy guarantees.

Differential privacy (DP)~\cite{dwork2006calibrating} has become a standard framework for analyzing privacy guarantees in learning-based systems. Informally, a randomized learning algorithm is differentially private if changing one individual training example has only a limited effect on the distribution of the algorithm's output. A central quantity in this setting is the sensitivity of the released statistic, which measures how much this statistic can change when a single training example is modified, added, or removed. In gradient-based learning, the quantity privatized is typically a minibatch gradient, or a function of it. Since neural-network gradients can have large or unbounded sensitivity, private training methods aim to control this sensitivity~\cite{dwork2014algorithmic}.

Methods such as DP-SGD~\cite{abadi2016deep} and DP-Adam~\cite{zhou2020private} enforce bounded sensitivity by clipping per-example gradients and adding calibrated Gaussian noise before the optimizer update. The resulting privacy loss is then tracked using accounting tools such as the Rényi differential privacy (RDP)~\cite{mironov2017renyi} accounting. While these methods are general and widely used for deterministic deep neural networks, their direct application to Bayesian deep learning raises additional challenges:  the interaction between injected privacy noise and the posterior sampling noise is non-trivial.

In this work, we take a step toward differentially private variational Bayesian deep learning by introducing \textsc{DP-IVON-Gradsq}, a private variant of the Improved Variational Online Newton (IVON)~\cite{shen2024variational}. Our method combines the posterior updates of IVON with the standard privacy mechanism of per-example gradient clipping, Gaussian noise injection, and RDP accounting. To reduce the interaction between variational sampling noise and privacy noise, we propose to employ the squared-gradient curvature estimation, where the curvature estimate is constructed from the privatized gradient using a noise-corrected squared-gradient approximation. We further develop an Opacus~\cite{yousefpour2021opacus}-compatible implementation comprising both a custom \textsc{DP-IVON-Gradsq} optimizer and a custom privacy-engine wrapper.   
Empirically, we evaluate the resulting privacy-utility trade-off and compare \textsc{DP-IVON-Gradsq} against standard differentially private optimizers such as DP-SGD and DP-Adam.

\section{Related Work}
\subsection{Differential Privacy for Bayesian Deep Learning}

A first line of work studies Bayesian neural networks trained by posterior sampling, in particular stochastic gradient Langevin dynamics (SGLD)~\cite{welling2011bayesian,li2016preconditioned}. SGLD can be interpreted as noisy SGD; consequently, DP-SGLD~\cite{wang2015privacy} can be viewed as a subclass of DP-SGD with an additional regularization term; for example, with a Gaussian prior, DP-SGLD becomes closely related to DP-SGD with an L2 norm penalty~\cite{zhang2023differentially}. This connection makes SGLD particularly appealing for private Bayesian learning. The “privacy for free” perspective~\cite{wang2015privacy} shows that, under bounded sensitivity, posterior sampling mechanisms can satisfy DP without requiring the same explicit clipping-and-noising procedure used in DP-SGD. However, in deep learning, the sensitivity of gradients or posterior updates is typically unbounded, so practical DP-SGLD variants often reintroduce clipping and Gaussian noise. This equivalence is useful conceptually, but it also means that DP-SGLD inherits many of the hyperparameter sensitivities of DP-SGD, while suffering from the absence of an analytic posterior and potentially high storage or sampling costs~\cite{zhang2023differentially}. 

A second line of work considers differentially private variational inference for Bayesian neural networks, including DP versions of Bayes by Backprop (DP-BBP)~\cite{blundell2015weight,zhang2023differentially}. Variational methods are attractive because they provide an explicit approximate posterior, often chosen as a diagonal Gaussian, which is more compact and scalable than storing many posterior samples. This makes DP-BBP more appealing than DP-SGLD from a memory perspective. However, private variational learning remains computationally demanding because the method combines stochastic variational optimization, Monte Carlo sampling, gradient clipping, and privacy noise~\cite{zhang2023differentially}. As a result, existing DP variational methods~\cite{zhang2023differentially} are slower and can exhibit weaker calibration than sampling-based alternatives such as DP-SGLD. This suggests that simply applying DP-SGD-style clipping and noising to variational objectives does not fully exploit the structure of Bayesian learning. 

A related approximation is Monte Carlo Dropout, which interprets dropout as approximate Bayesian inference and estimates predictive uncertainty through stochastic forward passes~\cite{gal2016dropout}. In a differentially private setting, DP-MC Dropout can be understood as training a dropout network with DP-SGD~\cite{zhang2023differentially}. This makes it relatively fast and easy to implement, while still providing a practical form of uncertainty quantification. However, its Bayesian interpretation is limited: the equivalence between dropout training and variational inference depends on specific assumptions, such as a Gaussian prior over weights, and does not generally hold for richer priors or posterior families. 

Overall, DP-SGLD benefits from a strong theoretical link between posterior sampling and privacy, but it is less scalable and lacks an analytic posterior. DP-BBP provides a compact variational posterior, but current private variants are computationally expensive and may suffer in calibration. DP-MC Dropout offers a simple and fast approximation, but its Bayesian interpretation is limited. These limitations motivate the development of new differentially private variational learning algorithms.

\subsection{Improved Variational Online Newton (IVON)}

Recent advances in scalable variational learning, particularly IVON~\cite{shen2024variational}, reopen the question of how to design private Bayesian optimizers that are both efficient and uncertainty-aware. Unlike standard variational inference methods~\cite{blei2017variational}, IVON, rooted in the Bayesian learning rule~\cite{khan2023bayesian}, embeds variational learning within the optimizer itself, without requiring changes to the network architecture or the training objective. It maintains a Gaussian posterior over the model weights using efficient second-order updates derived from reparameterized Hessian estimates, thereby enabling uncertainty-aware training at a computational cost close to that of deterministic Adam-based optimization~\cite{shen2024variational}. It learns both a posterior mean and a diagonal curvature-based uncertainty estimate, and it has been shown to scale to large models while improving calibration and predictive uncertainty. This makes IVON a promising candidate for differentially private Bayesian deep learning.

\section{Differentially Private Squared-Gradient IVON Variant}

\subsection{\textsc{DP-IVON-Gradsq} Algorithm}
\label{sec:dp-ivon-gradsq}

We propose a differentially private variant of IVON based on the squared-gradient curvature approximation, which we refer to as \textsc{DP-IVON-Gradsq}. The goal is to retain the variational-learning structure of IVON while enforcing DP through per-example gradient clipping and Gaussian noise injection. Unlike standard DP-SGD, which learns a deterministic point estimate, \textsc{DP-IVON-Gradsq} maintains a Gaussian posterior approximation over the model parameters. At each optimization step, the algorithm updates both the posterior mean and a diagonal curvature estimate, which in turn determines the posterior variance.

Let \(q(\theta)=\mathcal N(m,\operatorname{diag}(\sigma_v^2))\) denote the current variational posterior over model parameters, where \(m\) is the posterior mean and \(\sigma_v^2\) is the diagonal posterior variance. As in IVON, the posterior variance is parameterized through a diagonal curvature estimate \(h\) as
$$
\sigma_v^2 = \frac{1}{N_{\rm ess}(h+\delta_v)},
$$
where \(N_{\rm ess}\) is the effective sample size and \(\delta_v\) is a damping term, also playing the role of weight decay. At each iteration, we sample a parameter vector $\theta$ distributed as
$
\theta \sim \mathcal N(m,\operatorname{diag}(\sigma_v^2)),
$
compute gradients at the sampled parameters, and use them to update the variational posterior. 

To enforce DP, we sanitize the stochastic gradient before it is used by the IVON update. Given a minibatch \(B\) of size \(b\), we first compute per-example gradients
$
g_i = \nabla_\theta \mathcal L(\theta;x_i,y_i), (x_i,y_i)\in B.
$
Each gradient is then clipped using a global \(\ell_2\)-norm constraint:
$$
\bar g_i =
g_i
\min\left(
1,\frac{C}{\|g_i\|_2+\varepsilon_{\rm num}}
\right),
$$
where $C$ is a predefined clipping threshold and $\varepsilon_{\rm num}$ is a small numerical constant. The private averaged gradient is obtained by applying the Gaussian mechanism:
$$
g_{\rm dp}
=
\frac{1}{b}\sum_{i\in B}\bar g_i
+
\xi,
\qquad
\xi\sim
\mathcal N\left(
0,\frac{\sigma_p^2C^2}{b^2}I
\right),
$$
where $\sigma_p$ is the privacy noise multiplier. This corresponds to adding Gaussian noise with standard deviation $\sigma_p C$ to the clipped sum, followed by averaging over the minibatch.

A key design choice of our method is to reduce the interaction between the variational noise induced by posterior sampling and the Gaussian noise injected for privacy. To this end, we modify the Hessian approximation used in IVON. Instead of employing the original reparameterized estimator
$$
\widehat h_{\rm IVON}
=
g_{\rm dp}
\odot
\frac{\theta-m}{\sigma_v^2},
$$
which explicitly depends on the sampled perturbation $\theta-m$, we use a squared-gradient curvature estimate $g_{\rm dp}^2$. This choice avoids coupling the privacy noise in $g_{\rm dp}$ with the variational sampling noise in $\theta$, and makes the curvature estimate a post-processing of the privatized gradient.

However, because $g_{\rm dp}$ contains Gaussian privacy noise, using $g_{\rm dp}^2$ directly would add the variance of the injected noise to the curvature estimate. We therefore subtract the known privacy-noise variance, as a pre-projection correction for this additive contribution, and then project the result onto the nonnegative orthant to obtain a nonnegative squared-gradient curvature estimate:
\begin{align}
\label{eq:noiseCorrection}
\widehat v
=
\max\left(
g_{\rm dp}^2
-
\frac{\sigma_p^2C^2}{b^2},
0
\right).
\end{align}

The variance subtraction corrects the additive contribution of the Gaussian privacy noise before projection. However, the subsequent nonnegative projection introduces positive bias, while minibatch sampling and per-example clipping change the quantity being estimated. Therefore, $\widehat v$ should be interpreted as a stabilized, curvature proxy rather than an unbiased estimator.

All squares, divisions, and maximum operations are applied elementwise. The quantity \(\widehat v\) is computed only from the privatized gradient \(g_{\rm dp}\), and is therefore a post-processing operation; it does not consume any additional privacy budget. The corresponding curvature estimate is then given by
$$
\widehat h = N_{\rm ess}\widehat v.
$$
This scaling is inherited from IVON, where the curvature term corresponds to the Hessian approximation of the full objective rather than only the minibatch-averaged loss.

The diagonal curvature estimate is updated using the positivity-preserving IVON update:
\begin{align}
\label{eq:hessianUpdate}
h
\leftarrow
\beta_2 h
+
(1-\beta_2)\widehat h
+
\frac{(1-\beta_2)^2(h-\widehat h)^2}{2(h+\delta_v)}.    
\end{align}

The private gradient \(g_{\rm dp}\) is also used to update the gradient momentum:
\begin{align}
\label{eq:firstMoment}
s \leftarrow
\beta_1s+(1-\beta_1)g_{\rm dp},  
\end{align}
\begin{align}
\label{eq:firstMomentCorrected}
\bar s =
\frac{s}{1-\beta_1^t}.    
\end{align}

The posterior mean is then updated using the IVON natural-gradient-style preconditioning:
\begin{align}
    m
\leftarrow
m
-
\alpha
\frac{\bar s+\delta_vm}{h+\delta_v}.
\end{align}

The proposed procedure can be interpreted as inserting a DP gradient-sanitization layer (Algorithm~\ref{alg:dp-sanitize}) inside IVON as shown in Algorithm~\ref{alg:dp-ivon-gradsq}. The privacy mechanism acts only on the per-example gradients through clipping and Gaussian noise. The subsequent curvature estimate, momentum update, posterior mean update, and posterior variance update are post-processing operations of the privatized gradient and therefore do not increase the privacy loss. The overall privacy budget is tracked with a standard DP accountant using the sampling rate $q=b/|\mathcal D|$, the number of optimization steps $T$, the noise multiplier $\sigma_p$, and the target privacy parameter $\delta_p$. This yields a final privacy guarantee of the form $(\varepsilon,\delta_p)$-DP.

\begin{algorithm}[H]
\label{alg:dp-sanitize}
\small
\SetAlgoLined
\DontPrintSemicolon
\caption{\textsc{DPSanitizeGradsq}}

\KwIn{minibatch $B$, loss $\mathcal L$, posterior mean $m$, posterior variance $\sigma_v^2$, clipping norm $C$, noise multiplier $\sigma_p$}

\tcp{Variational weight sampling}
\[
\theta \sim \mathcal N(m,\operatorname{diag}(\sigma_v^2)).
\]

\tcp{Per-example gradient computation}
\[
g_i = \nabla_\theta \mathcal L(\theta;x_i,y_i),
\qquad (x_i,y_i)\in B.
\]

\tcp{Per-example gradient clipping}
\[
\bar g_i
=
g_i
\min\left(
1,
\frac{C}{\|g_i\|_2+\varepsilon_{\rm num}}
\right).
\]

\tcp{Private gradient aggregation}
\[
g_{\rm dp}
=
\frac{1}{b}\sum_{i\in B}\bar g_i
+
\mathcal N\left(
0,
\frac{\sigma_p^2C^2}{b^2}I
\right),
\qquad b=|B|.
\]

\tcp{Noise-corrected squared-gradient estimate}
\[
\widehat v
=
\max\left(
g_{\rm dp}^2
-
\frac{\sigma_p^2C^2}{b^2},
0
\right).
\]

\KwOut{private gradient $g_{\rm dp}$ and squared-gradient estimate $\widehat v$}

\end{algorithm}

\begin{algorithm}[H]
\small
\SetAlgoLined
\DontPrintSemicolon
\caption{\textsc{DP-IVON-Gradsq}}

\KwIn{dataset $\mathcal D$, loss $\mathcal L$, noise multiplier $\sigma_p$, privacy parameter $\delta_p$, clipping norm $C$, effective sample size $N_{\rm ess}$, learning rate $\alpha$, momenta $\beta_1,\beta_2$, damping $\delta_v$, initial mean $m_0$, initial curvature estimate $h_0$}

\BlankLine
\tcp{Initialization}
Initialize posterior mean $m \leftarrow m_0$\;
Initialize curvature estimate $h \leftarrow h_0$\;
Initialize gradient momentum $s \leftarrow 0$\;
Set posterior variance
$$
\sigma_v^2 \leftarrow \frac{1}{N_{\rm ess}(h+\delta_v)}.
$$

\BlankLine
\For{$t=1,\ldots,T$}{
    Sample minibatch $B \subset \mathcal D$\;

    \BlankLine
    \tcp{DP gradient sanitization (Algorithm~\ref{alg:dp-sanitize})}
    \[
    (g_{\rm dp},\widehat v)
    \leftarrow
    \textsc{DPSanitizeGradsq}
    \left(
    B,\mathcal L,m,\sigma_v^2,C,\sigma_p
    \right).
    \]

    \BlankLine
    \tcp{Squared-gradient curvature estimate}
    \[
    \widehat h
    \leftarrow
    N_{\rm ess}\widehat v.
    \]

    \BlankLine
    \tcp{Curvature estimate update}
    \[
    h
    \leftarrow
    \beta_2 h
    +
    (1-\beta_2)\widehat h
    +
    \frac{(1-\beta_2)^2(h-\widehat h)^2}{2(h+\delta_v)}.
    \]

    \BlankLine
    \tcp{Private gradient momentum update}
    \[
    s
    \leftarrow
    \beta_1s+(1-\beta_1)g_{\rm dp}. \quad 
    \bar s
    \leftarrow
    \frac{s}{1-\beta_1^t}.
    \]

    \BlankLine
    \tcp{Posterior mean update}
    \[
    m
    \leftarrow
    m
    -
    \alpha
    \frac{\bar s+\delta_vm}{h+\delta_v}.
    \]

    \BlankLine
    \tcp{Posterior variance update}
    \[
    \sigma_v^2
    \leftarrow
    \frac{1}{N_{\rm ess}(h+\delta_v)}.
    \]

    \BlankLine
    Update privacy accountant with sampling rate
    $
    q=\frac{|B|}{|\mathcal D|}.
    $
}

\BlankLine
\tcp{Final privacy budget}
Compute final privacy budget
\[
\varepsilon
=
\mathrm{Accountant}(q,T,\sigma_p,\delta_p).
\]

\KwOut{private posterior approximation $\mathcal N(m,\operatorname{diag}(\sigma_v^2))$ and privacy guarantee $(\varepsilon,\delta_p)$}
\label{alg:dp-ivon-gradsq}
\end{algorithm}

\subsection{Relationship to the DPAGD Framework}

Differentially Private Adaptive Gradient Descent (DPAGD)~\cite{zhou2020private} provides a general framework for private adaptive optimization. At each iteration, a noisy private gradient \(\tilde g_t\) is first computed by clipping per-example gradients and adding Gaussian noise. The optimizer then constructs two adaptive quantities from the history of privatized gradients:
$$
m_t = \phi_t(\tilde g_1,\ldots,\tilde g_t),
\qquad
v_t = \min(\psi_t(\tilde g_1,\ldots,\tilde g_t),\lambda) 
$$
where $\lambda$ is a positive hyperparameter. The optimizer then updates the parameters using a preconditioned gradient step

$$
\mathbf{\theta}_{t+1}
=
\mathbf{\theta}_t
-
\eta_t
\frac{m_t}{\sqrt{v_t}+\nu}.
$$

Different DP adaptive optimizers correspond to different choices of these two functions: for instance, DP-SGD uses the private gradient directly, while DP-Adam uses exponential moving averages of the private gradients and their squares~\cite{zhou2020private}. This formulation is useful for positioning our method, since \textsc{DP-IVON-Gradsq} can also be interpreted as belonging to this family at the level of private adaptive gradient construction. Indeed, after per-example clipping and Gaussian noise injection, the optimizer receives a privatized gradient $g_{\mathrm{dp},t}$, which is then used to update an Adam-like first-moment estimate (Equation \eqref{eq:firstMoment}).
The main difference lies in the adaptive second-order quantity. Instead of using the classical Adam-style second moment, \textsc{DP-IVON-Gradsq} builds a noise-corrected squared-gradient estimate (Equation \eqref{eq:noiseCorrection}),
which is then scaled by the effective sample size to obtain the IVON curvature target.
This target is used in the IVON Hessian recursion (Equation~\eqref{eq:hessianUpdate}).
Thus, \textsc{DP-IVON-Gradsq} can be viewed as a DP adaptive gradient method whose momentum function $\phi_t$ is the private gradient momentum (Equation~\eqref{eq:firstMomentCorrected})
$$\phi^{\textsc{DP-IVON-Gradsq}}_t = \bar{s}_t,$$
and whose adaptive function $$\psi^{\textsc{DP-IVON-Gradsq}}_t = h_t,$$ is the IVON squared-gradient curvature recursion (Equation~\eqref{eq:hessianUpdate}).

However, unlike DP-Adam or DP-RMSProp, the resulting update is Newton-like: the posterior mean is updated using $h_t+\delta_v$ directly as a curvature preconditioner, rather than using $\sqrt{v_t}+\nu$. This interpretation highlights that \textsc{DP-IVON-Gradsq} follows the DPAGD principle of adapting only from privatized gradients, while preserving the variational posterior structure of IVON.

\section{Implementation of \textsc{DP-IVON-Gradsq} with Opacus-Compatible Components}
The main implementation challenge is that Opacus~\cite{yousefpour2021opacus} does not natively support IVON’s variational posterior updates. We therefore reused selected Opacus components and implemented the IVON-specific components separately.

For the implementation of \textsc{DP-IVON-Gradsq}, we built upon the Opacus library while substantially extending its functionality to accommodate the Bayesian and variational components of IVON. Concretely, within our custom implementation of \texttt{make\_private}, the model is wrapped with Opacus's \texttt{GradSamp\allowbreak leModule}, which enables the computation of per-sample gradients during backpropagation and, consequently, the individual gradient clipping required for differential privacy. We also rely on Opacus's \texttt{RDPAccountant} to track the cumulative privacy loss across training steps and certify the final $(\varepsilon,\delta)$-DP guarantee.

The core optimization logic, however, was implemented from scratch to support the specific requirements of IVON in the differentially private setting. After per-sample clipping and gradient aggregation, calibrated Gaussian noise is added independently to each aggregated parameter-gradient tensor before the optimizer update. The resulting pipeline is encapsulated in a custom \texttt{DPOptimizer} whose interface and hook compatibility follow the structure of Opacus's \texttt{DPOptimizer}.

\begin{figure}[h]
    \centering
    \includegraphics[width=\linewidth]{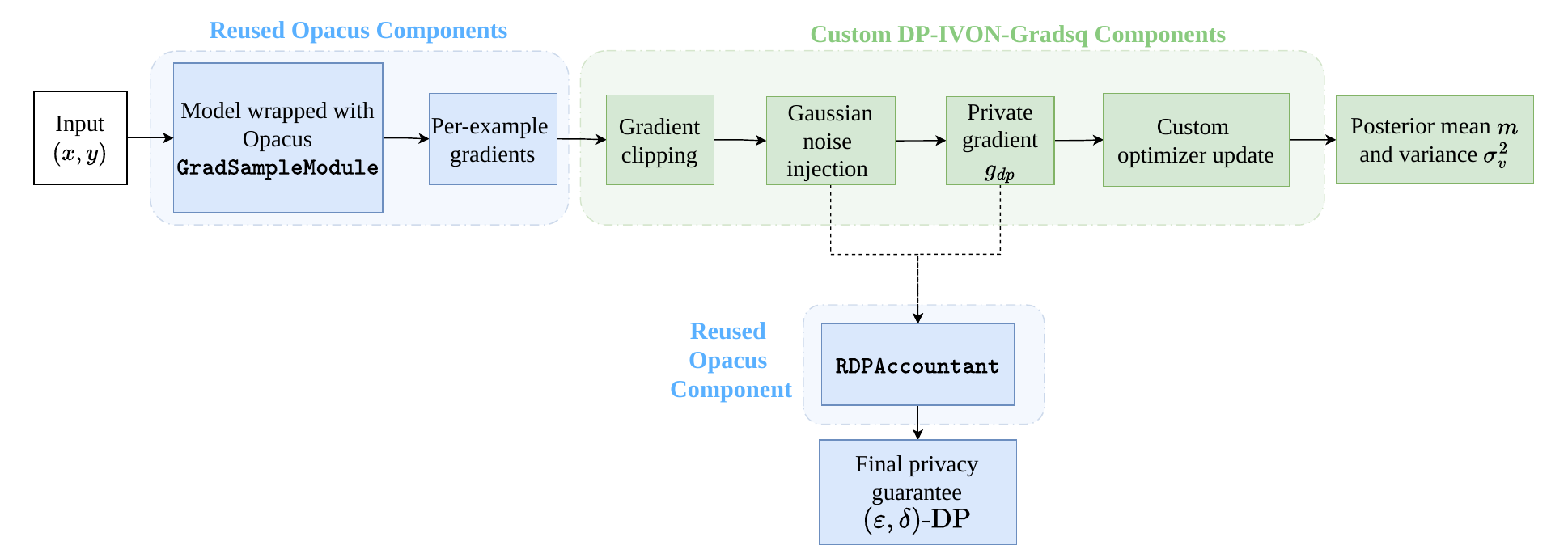}
    \caption{
Implementation pipeline of \textsc{DP-IVON-Gradsq}. We reuse Opacus components for per-example gradient computation through \texttt{GradSampleModule} and privacy accounting through \texttt{RDPAccountant}. The IVON-specific private gradient sanitization, squared-gradient curvature approximation, and variational posterior update are implemented in a custom optimizer compatible with the Opacus-style training interface.
}
    \label{fig:pipeline}
\end{figure}

\section{Experiments}

\subsection{Experimental setup}

We evaluate \textsc{DP-IVON-Gradsq} on CIFAR-10 using a convolutional neural network architecture and compare it against two standard private optimizers, DP-SGD and DP-Adam. 

\paragraph{Model Architecture.}
The model consists of three convolutional blocks followed by a two-layer classifier. We employ group normalization instead of batch normalization to make the architecture compatible with differentially private training.

\paragraph{Hyperparameters.}
All methods are trained for 150 epochs and evaluated over three random seeds in order to account for variability across runs. We perform a grid search over the main privacy and optimization hyperparameters, considering batch sizes in $\{128,256,512\}$, noise multipliers in $\{0.1,1.0,5.0,10.0,15.0\}$, privacy delta $\delta_p = 10^{-5}$, and clipping thresholds in $\{10,20\}$. The weight decay parameter is fixed to $10^{-4}$ for all experiments. For each optimizer, we first tune the learning rate over a grid of three candidate values and then use the best-performing value in the full comparison. The selected learning rates are $10^{-4}$ for DP-Adam, $10^{-2}$ for DP-SGD, and $10^{-1}$ for \textsc{DP-IVON-Gradsq}. 

\paragraph{Metrics.} 
In addition to accuracy, we report Expected Calibration Error (ECE) and Negative Log-Likelihood (NLL)  in order to evaluate not only the privacy-utility trade-off, but also the reliability of the predictive distributions. Accuracy measures the classification performance of the private models, while NLL penalizes both incorrect and overconfident predictions. ECE further evaluates whether predicted confidence scores are aligned with empirical correctness. We also report the final privacy budget $\varepsilon$ at a fixed target $\delta_p$, computed using RDP accounting. Since the same noise multiplier can correspond to different privacy guarantees depending on the batch size and number of training steps, we analyze performance as a function of $\varepsilon$ rather than only as a function of the noise scale.

\subsection{Results}

In all the following figures, the x-axis reports the final RDP-accounted privacy budget $\varepsilon$ and is shown on an inverted logarithmic scale. Larger values of $\varepsilon$ correspond to weaker privacy guarantees, while smaller values correspond to stronger privacy guarantees. Thus, curves should be read from left to right as moving from the low-privacy regime to the high-privacy regime.

\begin{figure}[h!]
    \centering
    \includegraphics[width=0.95\linewidth]{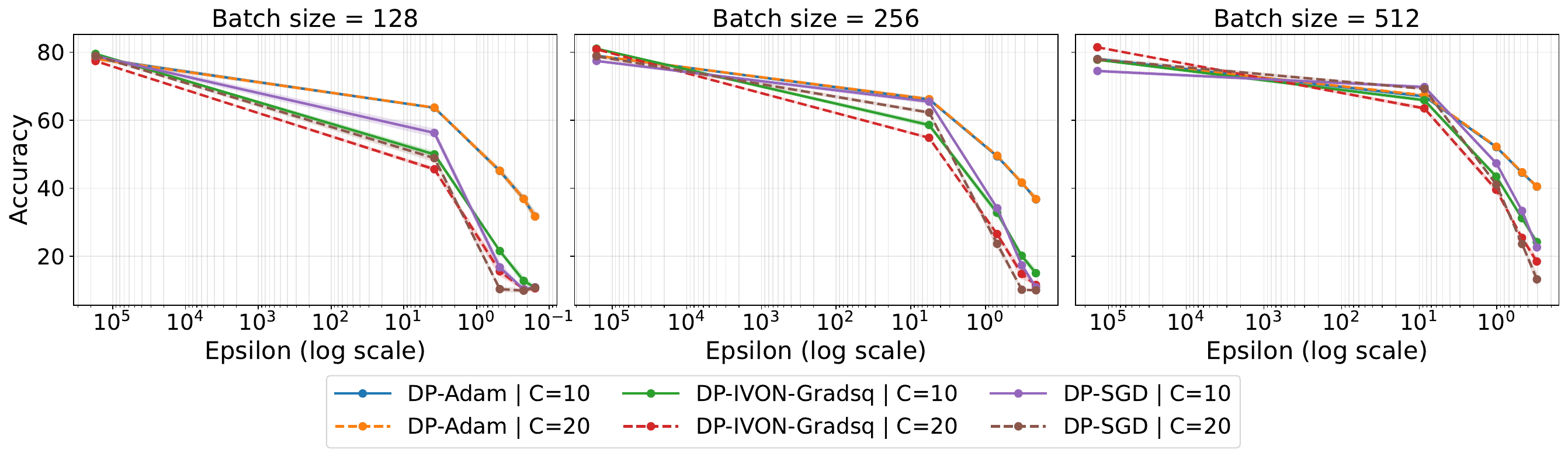}
    \caption{
    Privacy-accuracy trade-off on CIFAR-10 for DP-Adam, DP-IVON-Gradsq, and DP-SGD. 
    Test accuracy is reported as a function of the privacy budget $\varepsilon$ on an inverted logarithmic scale, for batch sizes $128$, $256$, and $512$. 
    Solid and dashed curves correspond to clipping norms $C=10$ and $C=20$, respectively. 
    }
    \label{fig:privacy-accuracy}
\end{figure}

\paragraph{Accuracy vs. Privacy.}
Figure~\ref{fig:privacy-accuracy} reports the accuracy-privacy trade-off of DP-IVON-Gradsq compared with DP-SGD and DP-Adam on CIFAR-10. The main observation is that DP-IVON-Gradsq is able to achieve competitive accuracy in the weak-privacy regime, where $\varepsilon$ is large, indicating that the proposed variant can achieve strong performance when the privacy perturbation is moderate. However, as $\varepsilon$ decreases, DP-IVON-Gradsq exhibits a sharper degradation than the strongest baseline, especially for smaller batch sizes. This behavior suggests that the variational update remains sensitive to the Gaussian noise introduced by the DP mechanism, even after replacing the original IVON Hessian approximation with the corrected squared-gradient curvature estimate.

Across the three batch-size settings, DP-IVON-Gradsq follows the expected behavior of DP gradient-based training ~\cite{raisa2024subsampling}. Larger batch sizes generally improve robustness in the medium-privacy regime and delay the point at which accuracy collapses. This is consistent with the standard DP-SGD trade-off: when Gaussian noise is added to the averaged clipped gradient, its scale is proportional to \(C/b\), so increasing the batch size reduces the effective perturbation applied to the private gradient. The clipping norm also affects this trade-off by controlling both the sensitivity of the gradient estimate and the magnitude of the injected noise. In this respect, DP-IVON-Gradsq exhibits the same qualitative dependence on batch size and clipping norm as standard DP optimizers. However, because the privatized gradient is also used to construct the curvature estimate, these hyperparameters influence not only the mean update but also the variational curvature estimate that determines the posterior variance.

\paragraph{NLL vs. Privacy.}

\begin{figure}[h!]
    \centering
    \includegraphics[width=0.95\linewidth]{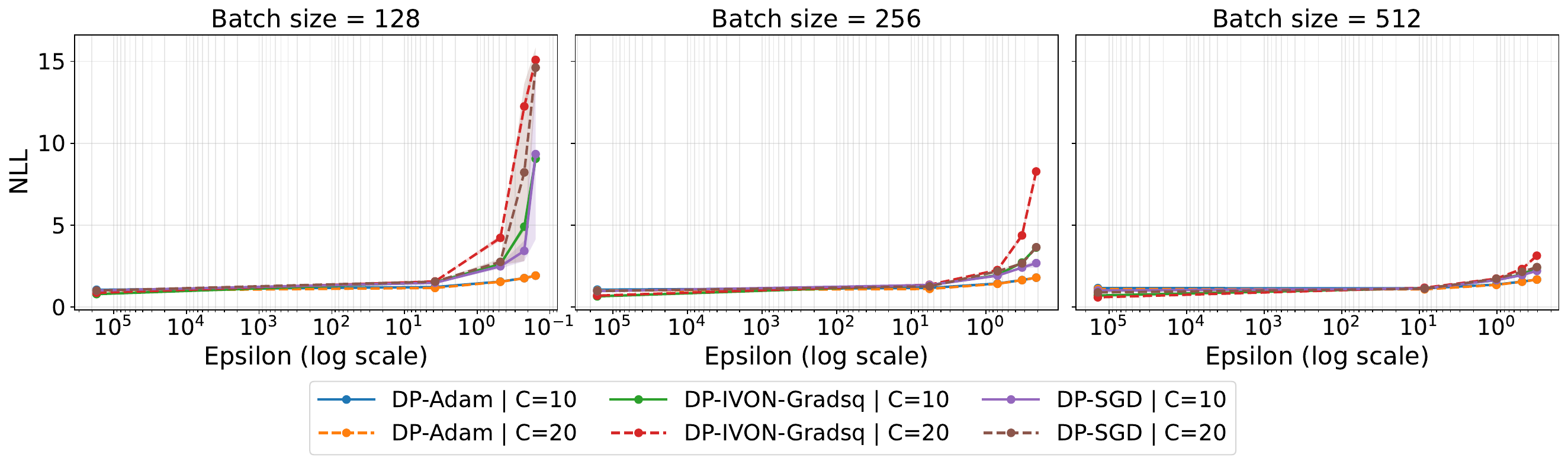}
    \caption{Privacy-NLL trade-off on CIFAR-10 for DP-Adam, DP-IVON-Gradsq, and DP-SGD. 
    Test NLL is reported as a function of the privacy budget $\varepsilon$ on an inverted logarithmic scale, for batch sizes $128$, $256$, and $512$. 
    Solid and dashed curves correspond to clipping norms $C=10$ and $C=20$, respectively.}
    \label{fig:privacy-nll}
\end{figure}

Figure~\ref{fig:privacy-nll} reports the evolution of test NLL as a function of the privacy budget $\varepsilon$. For DP-IVON-Gradsq, the results show that the method remains stable in the weak-privacy regime, where $\varepsilon$ is large, with NLL values better than or comparable to the DP-SGD and DP-Adam baselines. This indicates that, when the injected privacy noise is moderate, the proposed variant can preserve the uncertainty-aware behavior of IVON while satisfying DP.

However, DP-IVON-Gradsq becomes more sensitive as the privacy constraint becomes stronger. In particular, for smaller batch sizes, the NLL increases sharply at low $\varepsilon$, especially when using the larger clipping norm $C=20$. This degradation is stronger than what is observed for DP-SGD, suggesting that the privatized gradient affects DP-IVON-Gradsq through two coupled mechanisms:
\begin{itemize}
    \item It directly drives the posterior mean update.
    \item It also determines the curvature estimate through a noise-corrected squared-gradient approximation, as in Equation~\eqref{eq:noiseCorrection}, which may lead to less reliable posterior variance estimates and overconfident or poorly calibrated predictions in the high-privacy regime.
\end{itemize}

The batch-size effect follows the standard trend observed for accuracy. Larger batch sizes reduce the effective noise scale on the averaged clipped gradient and therefore improve stability across all optimizers. For DP-IVON-Gradsq, this effect is especially visible when moving from batch size \(128\) to \(512\), where the NLL growth under stronger privacy constraints becomes substantially less severe. This suggests that DP-IVON-Gradsq benefits from larger minibatches, not only for improving the private gradient estimate, but also for stabilizing the curvature estimate used in the variational update.

\begin{figure}[h]
    \centering
    \includegraphics[width=0.95\linewidth]{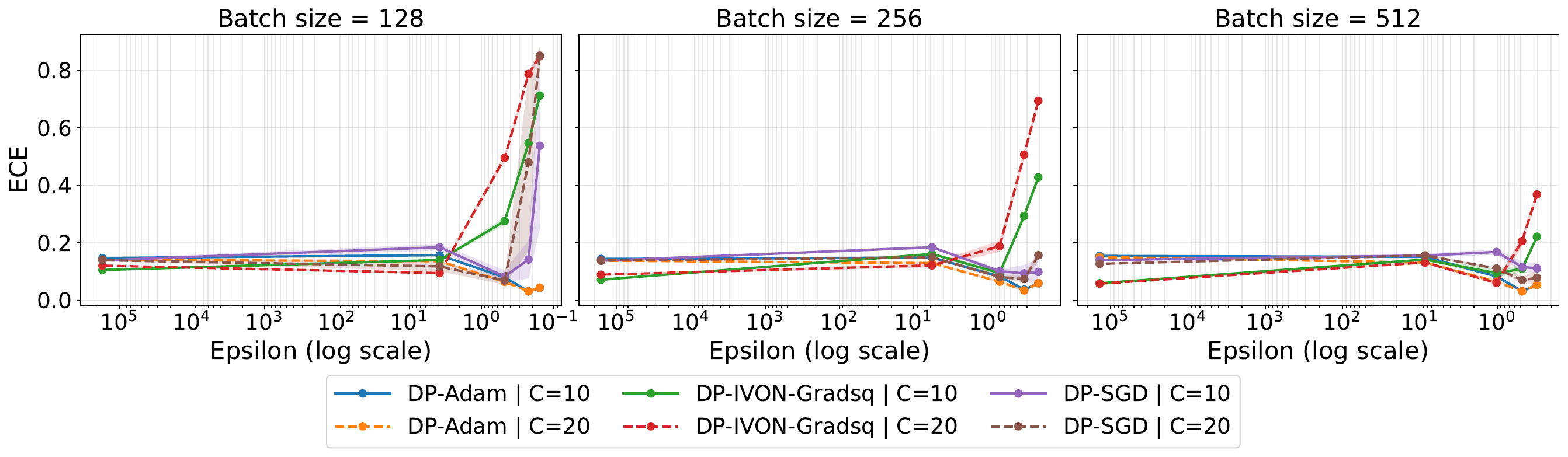}
    \caption{Privacy-calibration trade-off on CIFAR-10 for DP-Adam, DP-IVON-Gradsq, and DP-SGD. 
    Test ECE is reported as a function of the privacy budget $\varepsilon$ on an inverted logarithmic scale, for batch sizes $128$, $256$, and $512$. 
    Solid and dashed curves correspond to clipping norms $C=10$ and $C=20$, respectively. }
    \label{fig:privacy-ece}
\end{figure}

\paragraph{ECE vs. Privacy.}
Figure~\ref{fig:privacy-ece} reports the calibration behavior of the private optimizers through ECE. DP-IVON-Gradsq appears best calibrated in the weak and medium privacy regimes, where $\varepsilon$ is large or moderate. However, as the privacy budget becomes smaller, DP-IVON-Gradsq exhibits a sharper increase in ECE, especially for $C=20$ and smaller batch sizes. 

This indicates that, in the high-privacy regime, the injected Gaussian noise affects not only classification accuracy but also the reliability of the predicted confidence scores. The effect is particularly visible for batch sizes $128$ and $256$, where DP-IVON-Gradsq becomes substantially less calibrated at low $\varepsilon$. This behavior is consistent with the NLL results and suggests that the curvature estimate is sensitive to strong privacy perturbations.

\section{Discussion}
The empirical results show that \textsc{DP-IVON-Gradsq} is feasible as a private variational optimizer and can remain competitive when the privacy constraint is not too stringent, without introducing an additional computational cost specific to variational learning. Nevertheless, its faster degradation in the high-privacy regime highlights a central challenge of differentially private variational learning. We hypothesize that this behavior is partly caused by the noise-correction and projection step used in the squared-gradient estimator (Equation~\eqref{eq:noiseCorrection}). In particular, when gradients become small toward the end of training, subtracting the privacy-noise variance can dominate the squared-gradient signal, causing the projected squared-gradient estimate $\widehat v$ to be zero for many coordinates. In this case, the corresponding curvature target $\widehat h = N_{\rm ess}\widehat v$ also becomes zero, and the curvature update no longer receives an informative curvature signal.
The running curvature estimate then mainly decays slowly through the momentum update. This can weaken the interpretation of $h$ as a curvature estimate and lead to poorly informative posterior variance estimates.

This motivates further work on more robust private curvature estimation and better tuning strategies for IVON-specific hyperparameters under strong privacy constraints.

\section{Conclusion and Future Work}

In this work, we investigated the integration of differential privacy with variational Bayesian deep learning through IVON, with computational cost close to deterministic DP optimizers. We introduced \textsc{DP-IVON-Gradsq}, a differentially private variant of IVON. The proposed variant was designed to reduce the direct interaction between posterior-sampling noise and privacy noise by constructing the curvature estimate from the privatized gradient through a noise-corrected squared-gradient estimator.

Our findings highlight both the promise and the difficulty of differentially private variational optimization. While \textsc{DP-IVON-Gradsq} provides a practical route toward uncertainty-aware private training, the results also show that more robust private curvature estimation is needed for high-privacy regimes. Future work will investigate alternative curvature estimators and stronger theoretical analyses of the interaction between privacy noise, variational sampling, and posterior uncertainty. 

\section*{Acknowledgments}
This work was supported by the IMT “Futur, Ruptures \& Impacts” programme, by the European Research Council (ERC) under the European Union’s Horizon 2020 research and innovation programme (Grant Agreement No. 101003431, SONATA), and by the Smart Networks and Services Joint Undertaking (SNS JU) under the European Union’s Horizon Europe research and innovation programme (Grant Agreement No. 101192080, 6G-LEADER).

\bibliographystyle{splncs04}
\bibliography{ref}

@inproceedings{dwork2006calibrating,
  title={Calibrating {N}oise to {S}ensitivity in {P}rivate {D}ata {A}nalysis},
  author={Dwork, Cynthia and McSherry, Frank and Nissim, Kobbi and Smith, Adam},
  booktitle={Theory of Cryptography Conference},
  pages={265--284},
  year={2006},
  organization={Springer}
}

@inproceedings{abadi2016deep,
  title={Deep {L}earning with {D}ifferential {P}rivacy},
  author={Abadi, Martin and Chu, Andy and Goodfellow, Ian and McMahan, H Brendan and Mironov, Ilya and Talwar, Kunal and Zhang, Li},
  booktitle={Proceedings of the 2016 ACM SIGSAC Conference on Computer and Communications Security},
  pages={308--318},
  year={2016}
}

@inproceedings{mironov2017renyi,
  title={R{\'e}nyi {D}ifferential {P}rivacy},
  author={Mironov, Ilya},
  booktitle={2017 IEEE 30th Computer Security Foundations Symposium (CSF)},
  pages={263--275},
  year={2017},
  organization={IEEE}
}

@article{dwork2014algorithmic,
  title={The {A}lgorithmic {F}oundations of {D}ifferential {P}rivacy},
  author={Dwork, Cynthia and Roth, Aaron},
  journal={Foundations and Trends{\textregistered} in Theoretical Computer Science},
  volume={9},
  number={3-4},
  pages={211--487},
  year={2014},
  publisher={Emerald Publishing Limited}
}

@InProceedings{shen2024variational,
  title = 	 {Variational {L}earning is {E}ffective for {L}arge {D}eep {N}etworks},
  author =       {Shen, Yuesong and Daheim, Nico and Cong, Bai and Nickl, Peter and Marconi, Gian Maria and Raoul, Bazan Clement Emile Marcel and Yokota, Rio and Gurevych, Iryna and Cremers, Daniel and Khan, Mohammad Emtiyaz and M\"{o}llenhoff, Thomas},
  booktitle = 	 {Proceedings of the 41st ICML},
  pages = 	 {44665--44686},
  year = 	 {2024},
  volume = 	 {235},
  series = 	 {PMLR},
  publisher =    {PMLR}
}

@article{yousefpour2021opacus,
  author       = {Ashkan Yousefpour and
                  Igor Shilov and
                  Alexandre Sablayrolles and
                  Davide Testuggine and
                  Karthik Prasad and
                  Mani Malek and
                  John Nguyen and
                  Sayan Ghosh and
                  Akash Bharadwaj and
                  Jessica Zhao and
                  Graham Cormode and
                  Ilya Mironov},
  title        = {Opacus: {U}ser-{F}riendly {D}ifferential {P}rivacy {L}ibrary in {P}y{T}orch},
  journal      = {CoRR},
  year         = {2021}
}

@inproceedings{welling2011bayesian,
  title={Bayesian {L}earning via {S}tochastic {G}radient {L}angevin {D}ynamics},
  author={Welling, Max and Teh, Yee W},
  booktitle={Proceedings of the 28th ICML},
  pages={681--688},
  year={2011}
}

@inproceedings{li2016preconditioned,
  title={Preconditioned {S}tochastic {G}radient {L}angevin {D}ynamics for {D}eep {N}eural {N}etworks},
  author={Li, Chunyuan and Chen, Changyou and Carlson, David and Carin, Lawrence},
  booktitle={Proceedings of the AAAI Conference on Artificial Intelligence},
  volume={30},
  number={1},
  year={2016}
}

@inproceedings{zhang2023differentially,
  title={Differentially {P}rivate {B}ayesian {N}eural {N}etworks on {A}ccuracy, {P}rivacy and {R}eliability},
  author={Zhang, Qiyiwen and Bu, Zhiqi and Chen, Kan and Long, Qi},
  booktitle={Joint European Conference on Machine Learning and Knowledge Discovery in Databases},
  pages={604--619},
  year={2023},
  organization={Springer}
}

@inproceedings{wang2015privacy,
  title={Privacy for {F}ree: {P}osterior {S}ampling and {S}tochastic {G}radient {M}onte {C}arlo},
  author={Wang, Yu-Xiang and Fienberg, Stephen and Smola, Alex},
  booktitle={ICML},
  pages={2493--2502},
  year={2015},
  organization={PMLR}
}

@inproceedings{gal2016dropout,
  title={Dropout as a {B}ayesian {A}pproximation: {R}epresenting {M}odel {U}ncertainty in {D}eep {L}earning},
  author={Gal, Yarin and Ghahramani, Zoubin},
  booktitle={ICML},
  pages={1050--1059},
  year={2016},
  organization={PMLR}
}

@inproceedings{blundell2015weight,
  title={Weight {U}ncertainty in {N}eural {N}etworks},
  author={Blundell, Charles and Cornebise, Julien and Kavukcuoglu, Koray and Wierstra, Daan},
  booktitle={ICML},
  pages={1613--1622},
  year={2015},
  organization={PMLR}
}

@article{blei2017variational,
  title={Variational {I}nference: {A} {R}eview for {S}tatisticians},
  author={Blei, David M and Kucukelbir, Alp and McAuliffe, Jon D},
  journal={{Journal of the American Statistical Association}},
  volume={112},
  number={518},
  pages={859--877},
  year={2017},
  publisher={Taylor \& Francis}
}

@article{khan2023bayesian,
  title={The {B}ayesian {L}earning {R}ule},
  author={Khan, Mohammad Emtiyaz and Rue, H{\aa}vard},
  journal={JMLR},
  volume={24},
  number={281},
  pages={1--46},
  year={2023}
}

@article{zhou2020private,
  title={Private {S}tochastic {N}on-{C}onvex {O}ptimization: {A}daptive {A}lgorithms and {T}ighter {G}eneralization {B}ounds},
  author={Zhou, Yingxue and Chen, Xiangyi and Hong, Mingyi and Wu, Zhiwei Steven and Banerjee, Arindam},
  journal={arXiv preprint arXiv:2006.13501},
  year={2020}
}

@InProceedings{raisa2024subsampling,
  title = 	 {Subsampling is not {M}agic: {W}hy {L}arge {B}atch {S}izes {W}ork for {D}ifferentially {P}rivate {S}tochastic {O}ptimisation},
  author =       {R\"{a}is\"{a}, Ossi and J\"{a}lk\"{o}, Joonas and Honkela, Antti},
  booktitle = 	 {Proceedings of the 41st ICML},
  pages = 	 {41959--41981},
  year = 	 {2024},
  volume = 	 {235},
  series = 	 {PMLR},
  month = 	 {21--27 Jul},
  publisher =    {PMLR}
}

\end{document}